\documentclass[10pt,twocolumn,letterpaper]{article}

\usepackage{cvpr}     

\usepackage[dvipsnames]{xcolor}

\definecolor{cvprblue}{rgb}{0.21,0.49,0.74}
\usepackage[pagebackref,breaklinks,colorlinks,citecolor=cvprblue]{hyperref}
\usepackage{pifont}
\usepackage{multirow}
\usepackage{makecell}
\usepackage{diagbox}
\usepackage{colortbl}
\usepackage{algorithm}
\usepackage{algpseudocode}
\usepackage[most]{tcolorbox}
\def\confName{CVPR}
\def\confYear{2026}

\author{
Yi Xin\textsuperscript{\rm 1, 2, 3 *}, 
Siqi Luo\textsuperscript{\rm 3, 4 *}, 
Tianxiang Xu\textsuperscript{\rm 5},
Qi Qin\textsuperscript{\rm 3}, 
Haoxing Chen\textsuperscript{\rm 1},
Kaiwen Zhu\textsuperscript{\rm 3, 4}, 
Zhiwei Zhang\textsuperscript{\rm 2}, \\
Yangfan He\textsuperscript{\rm 2}, 
Rongchao Zhang\textsuperscript{\rm 5},
Jinbin Bai\textsuperscript{\rm 6},
Shuo Cao\textsuperscript{\rm 3}, 
Bin Fu\textsuperscript{\rm 3},
Junjun He\textsuperscript{\rm 3}, 
Yihao Liu\textsuperscript{\rm 3},\\
Yuewen Cao\textsuperscript{\rm 3}\footnotemark[2]~,~
Xiaohong Liu\textsuperscript{\rm 2, 4}\footnotemark[2]
\\
\textsuperscript{\rm 1}{Nanjing University},~
\textsuperscript{\rm 2}{Shanghai Innovation Institute},
~\textsuperscript{\rm 3}{Shanghai AI Lab}, \\ \textsuperscript{\rm 4}{Shanghai Jiao Tong University},~
\textsuperscript{\rm 5}{Peking University},~
\textsuperscript{\rm 6}{National University of Singapore}
\\
\raisebox{-0.3em}{\includegraphics[height=1.44em]{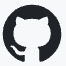}} \textbf{Code:}  \url{https://github.com/Alpha-VLLM/Lumina-DiMOO}
}

\title{dMLLM-TTS: \\Self-Verified and Efficient Test-Time Scaling for Diffusion Multi-Modal \\Large Language Models}

\begin{document}
\twocolumn[{
 \renewcommand\twocolumn[1][]{#1}
\maketitle
 \thispagestyle{empty}
 \pagestyle{empty}
 \begin{center}
     \captionsetup{type=figure}
    \vspace{-2.2em} 
    \includegraphics[width=1\linewidth]{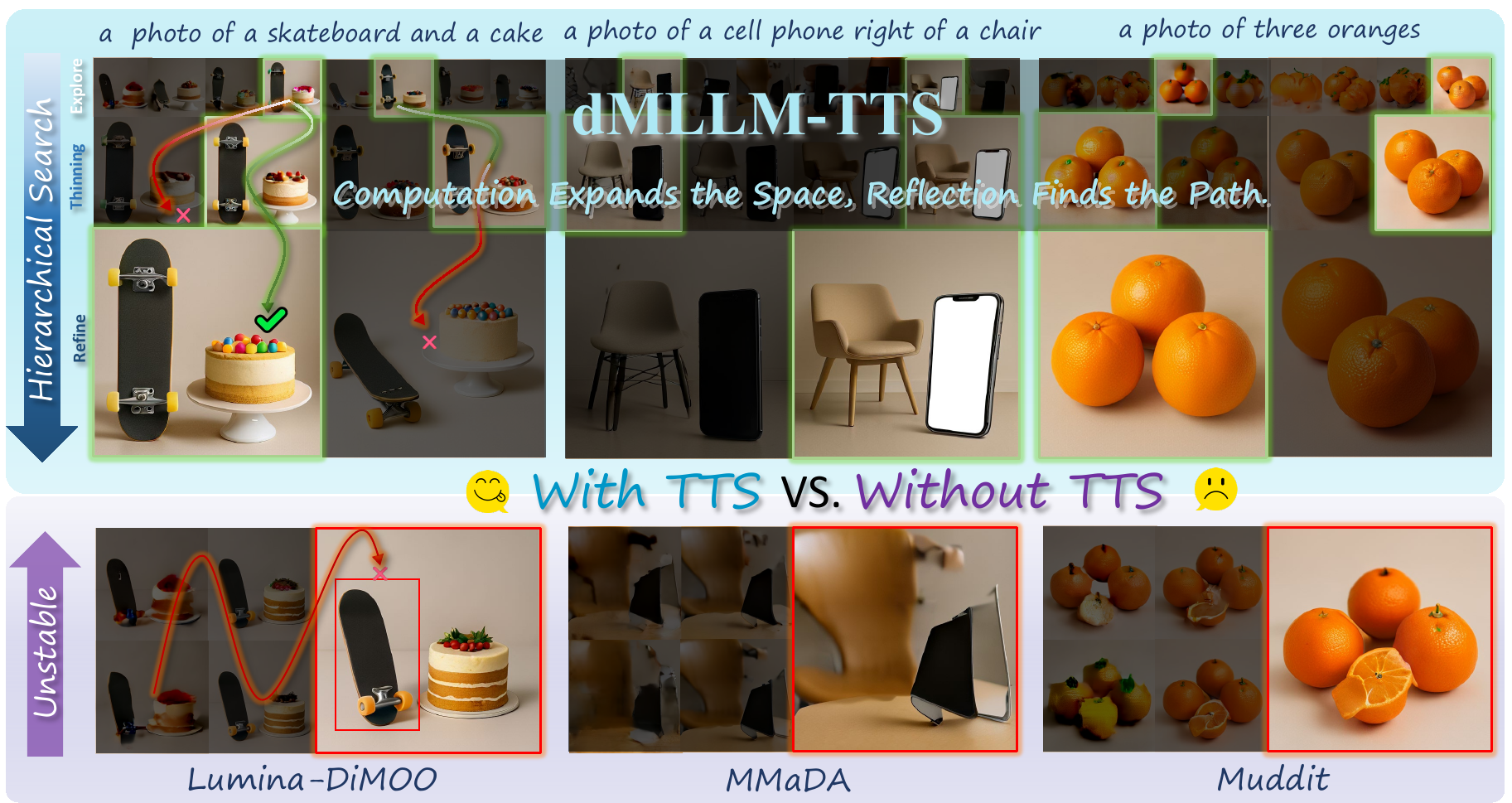}
    \vspace{-1.35em}
     \captionof{figure}{\textbf{dMLLM-TTS:} We present the generative effects and performance improvements achieved by applying Test-Time Scaling (TTS) to dMLLMs. Images generated with TTS exhibit higher quality and stronger prompt alignment than those generated without TTS.}
     \label{fig:teaser}
     \vspace{0.1cm}
 \end{center}
}]

\let\tempfootnote\thefootnote
\let\thefootnote\relax
\footnotetext{$^*$ Equal Contribution}
\footnotetext{$\dagger$ Corresponding Author}

\begin{abstract}
    Diffusion Multi-modal Large Language Models (dMLLMs) have recently emerged as a novel architecture unifying image generation and understanding. However, developing effective and efficient Test-Time Scaling (TTS) methods to unlock their full generative potential remains an underexplored challenge.
    To address this, we propose dMLLM-TTS, a novel framework operating on two complementary scaling axes: (1) trajectory exploration scaling to enhance the diversity of generated hypotheses, and (2) iterative refinement scaling for stable generation. 
    Conventional TTS approaches typically perform linear search across these two dimensions, incurring substantial computational costs of 
    $\mathcal{O}(NT)$ and requiring an external verifier for best-of-N selection.
    To overcome these limitations, we propose two innovations. First, we design an efficient hierarchical search algorithm with $\mathcal{O}(N+T)$ complexity that adaptively expands and prunes sampling trajectories. 
    Second, we introduce a self-verified feedback mechanism that leverages the dMLLMs' intrinsic image understanding capabilities to assess text–image alignment, eliminating the need for external verifier.
    Extensive experiments on the GenEval benchmark across three representative dMLLMs (e.g., Lumina-DiMOO, MMaDA, Muddit) show that our framework substantially improves generation quality while achieving up to 6$\times$ greater efficiency than linear search.
\end{abstract}    
\vspace{-0.2in}
\section{Introduction}
\label{sec:intro}
Recent advances in text-to-image (T2I) generation, driven by powerful diffusion models~\cite{chen2023pixart,qin2025lumina,podell2023sdxl,flux2024} and autoregressive (AR) models~\cite{xin2025lumina,chen2025januspro}, have achieved significant breakthroughs.
This success is largely attributed to the scalability of training resources, as evidenced by consistent performance improvements with increased training compute, model parameters, and dataset sizes.
However, this training-time scaling paradigm is encountering diminishing returns, constrained by exorbitant computational costs and a growing scarcity of high-quality data.
These challenges have sparked interest in test-time scaling (TTS), a more efficient strategy that enhances the capabilities of pretrained T2I generative models by allocating additional computational resources during inference, which has also yielded highly satisfactory results.

Most recently, Diffusion Multi-Modal Large Language Models (dMLLMs)~\cite{yang2025mmada,xin2025luminadimoo,shi2025muddit} have emerged, representing a fundamental architectural evolution. 
dMLLMs feature a unified architecture that natively integrates both image generation and understanding within a discrete diffusion modeling. 
This development has motivated researchers to investigate test-time scaling strategies specifically for dMLLMs.

In this paper, we introduce dMLLM-TTS, the first test-time scaling framework for dMLLMs that harmonizes the scaling strategy, verification mechanism, and search algorithm into a single, elegant inference process.
Drawing inspiration from TTS principles in diffusion models~\cite{xie2025sana,ma2025inference,singhal2025general,zhuo2025reflection}, we conceptualize test-time scaling in dMLLMs along two complementary axes:
(1) \textbf{Trajectory Exploration Scaling}, which broadens the hypothesis space by generating $N$ diverse initial samples to increase the likelihood of finding images that match the text prompt, 
and (2) \textbf{Iterative Refinement Scaling}, which deepens the generation process with $T$ refinement steps per trajectory to enhance generative stability and final image quality. 
Combining these two dimensions can significantly enhance the generative capabilities of dMLLMs.

Furthermore, following the experience of TTS in diffusion models, a linear search is typically performed along these two dimensions and an external Vision-Language Model (VLM)~\cite{xie2025sana,jaech2024openai,hessel2021clipscore} is employed to verify and score the final candidates, selecting the one that best matches the prompt. 
However, this approach has a sampling complexity of $\mathcal{O}(NT)$, and deploying an external VLM incurs additional overhead. 
This leads us to two fundamental questions: (1) \textit{Can a dMLLM verify its own generated images, thereby eliminating the need for an external model?} and (2) \textit{Can we design an efficient search algorithm that adaptively allocates compute to the most promising generative trajectories?}

This paper answers both questions affirmatively. 
We introduce a Self-Verified Feedback (SVF) mechanism that repurposes the dMLLM’s intrinsic multimodal understanding to assess text–image alignment. 
Specifically, we frame this as a question-answering task, querying the model on whether a given text prompt accurately describes the generated image. 
The probability of a “yes” response is then used as the alignment score.
Guided by this feedback, we design a Hierarchical Trajectory Search (HTS) algorithm. 
HTS implements a coarse-to-fine strategy that first broadly explores diverse structural hypotheses, then prunes low-potential trajectories, and finally refines the surviving high-potential trajectories for enhanced detail.
The SVF score is instrumental in both pruning unpromising trajectories and selecting the best sample from the final candidates. This intelligent allocation reduces the search complexity to a near-linear $\mathcal{O}(N+T)$.

We conduct extensive experiments using the GenEval~\cite{ghosh2024geneval} benchmark across three representative dMLLMs, (\textit{e.g.}, Lumina-DiMOO~\cite{xin2025luminadimoo}, MMaDA~\cite{yang2025mmada}, Muddit~\cite{shi2025muddit}), consistently observing significant performance improvements.
Furthermore, our method surpasses the linear search TTS baseline, offering higher efficiency and delivering superior final outputs.
These results underscore the potential of dMLLM-TTS to enhance image generation quality without the need for additional large-scale training.

Our main contributions are summarized as follows:
\begin{itemize}
    \item \textbf{Pioneering TTS Framework.} We establish the first test-time scaling framework for dMLLMs that integrates scaling strategy, verification, and search algorithm. 
    
    \item \textbf{Novel Verifier.} We introduce a self-verification mechanism that leverages the model’s inherent image understanding to internally evaluate generative outcomes, eliminating external verifiers.

    \item \textbf{Efficient Search Algorithm.} We present the Hierarchical Trajectory Search, which optimizes efficiency, achieving $\mathcal{O}(N+T)$ complexity, outperforming conventional linear search baseline with $\mathcal{O}(NT)$ complexity.

    \item \textbf{Superior Performance.} The proposed TTS framework elevates dMLLMs to match state-of-the-art generation models, significantly boosting image quality.
\end{itemize}

\begin{figure*}[!t]
    \centering
    \vspace{-0.18in}
    \includegraphics[width=1.0\linewidth]{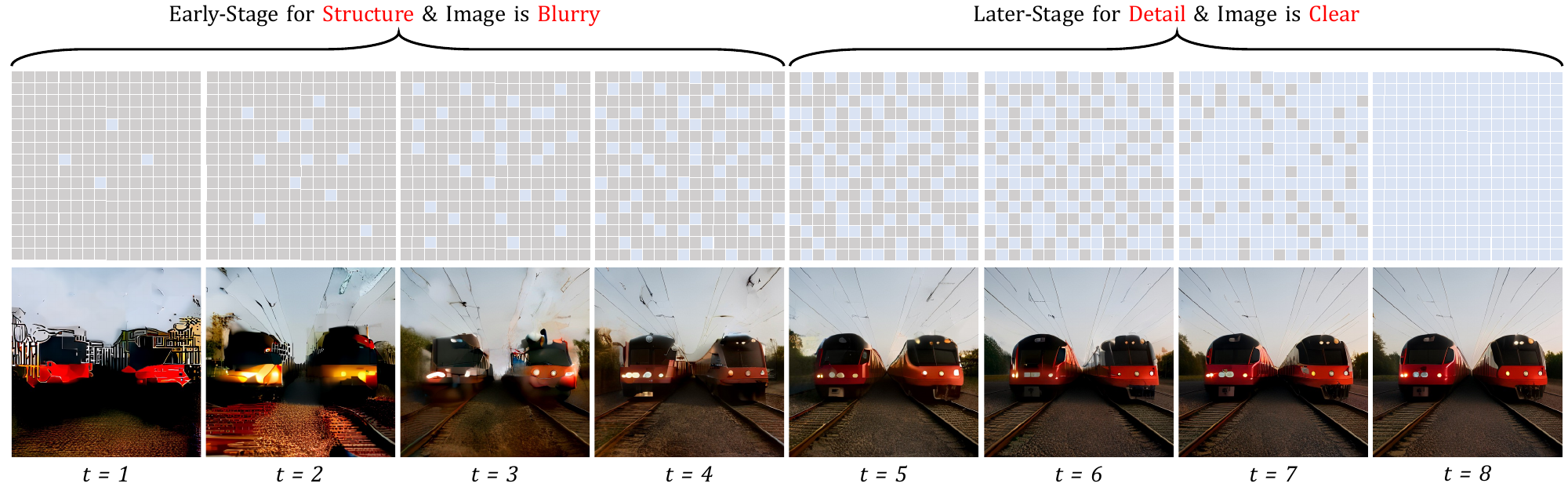}
    \vspace{-0.25in}
    \caption{\textbf{Visualization of the image generation process in dMLLMs.} The first row shows the input latent masks at each step, and the second row depicts the corresponding outputs. Sampling begins with fully masked tokens (gray) and gradually fills the discrete multimodal token space with increasingly confident predictions (blue).}
    \label{fig:genprocess}
    \vspace{-0.1in}
\end{figure*}
\section{Related Work}
\label{sec:related}
\noindent {\textbf{Diffusion Multi-Modal Large Language Models (dMLLMs).} 
Diffusion modeling has recently progressed from continuous visual generation to discrete multi-modal generation. 
Early discrete diffusion methods~\cite{austin2021d3pm,lou2023sedd} established the foundation for token-level denoising in discrete spaces. 
Building on these formulations, diffusion large language models (dLLMs) ~\cite{gong2024scalingdlm,nie2025llada,llada1.5} have demonstrated that language generation can be reimagined as an iterative parallel denoising process instead of following the traditional autoregressive approach. 
Extending this paradigm to multi-modal domains, LLaDA-V~\cite{lladav} introduced visual instruction tuning, showcasing the potential of diffusion-based language models to seamlessly connect textual and visual modalities. 
Subsequent research further explores this direction toward unified diffusion multi-modal large language models (dMLLMs)~\cite{yang2025mmada,shi2025muddit,xin2025luminadimoo}, which can \textit{perform both multi-modal generation and understanding}.

\noindent {\textbf{Test-Time Scaling for Image Generation.}
Prior to the emergence of dMLLMs, diffusion models~\cite{chen2023pixart,gaolumina,zhuolumina,qin2025lumina,xie2024sana,yi2024towards} and autoregressive models~\cite{liu2024lumina,xin2025lumina,xin2025resurrect,wang2024emu3,chen2025januspro} dominated as the primary paradigms for image generation.
Recent studies on these architectures have offered valuable insights into the principles of TTS.
For diffusion models, scaling typically expands the continuous noise space~\cite{xie2025sana,ma2025inference,singhal2025general} or increases the number of sampling steps~\cite{ma2025inference,singhal2025general} to improve quality and diversity, with further progress driven by reflection-based optimization~\cite{zhuo2025reflection,li2025reflect}.
Autoregressive models, on the other hand, allow for a distinct type of scaling by directly operating over discrete token sequences~\cite{xin2025lumina,chen2025tts}. 
Previous TTS approaches commonly rely on external verifiers~\cite{hessel2021clipscore,xu2024imagereward,hurst2024gpt} to evaluate generated images, 
yet such designs often require additional VLM deployments.
This motivates exploring TTS within dMLLMs, which possess \textit{unified multimodal understanding and generation capabilities that naturally support in-loop self-verification}.
\vspace{-0.05in}
\section{Methodology}
\label{sec:method}
\subsection{Preliminary}
\paragraph{Image Generation Process of dMLLMs.} dMLLMs generate images via a parallel, diffusion-based process that iteratively denoises a fully masked sequence into the target output, as shown in Figure~\ref{fig:genprocess}. 
Let $\mathcal{Y}$ be the token vocabulary and \texttt{[Mask]} $\in\mathcal{Y}$ the special mask token. 
Given a text prompt $C$, the dMLLMs generate an image token sequence $Z =(z_1, z_2, ...., z_M)$ (a total of M tokens) through $T$ discrete denoising steps, indexed by $t = 1$ up to $T$. 
The initial state $Z_0$ is a fully masked sequence:
\begin{equation}
    Z_0 = (\underbrace{\texttt{[Mask]}, \texttt{[Mask]}, ... ,\texttt{[Mask]}, \texttt{[Mask]}}_{M}).
\end{equation}

At each step $t$, dMLLMs predicts the tokens for all the masked locations:
\begin{equation}
    Z_t = (z_1^t, ... z_{M-1}^t, z_M^t).
\end{equation}
Subsequently, tokens with high confidence is the final choice and low confidence are re-masked and used as input for prediction at step $t+1$:
\vspace{-0.05in}
\begin{equation}
    Z_t = (\texttt{[Mask]}, z_2^t, ... ,\texttt{[Mask]}, z_M^t).
\end{equation}

\vspace{-0.05in}
To decode a generated token sequence $Z$ into an image, a specific discrete tokenizer~\cite{xie2024show,patil2024amused} is essential.

\begin{figure*}[!t]
    \centering
    \vspace{-0.1in}
    \includegraphics[width=1.01\linewidth]{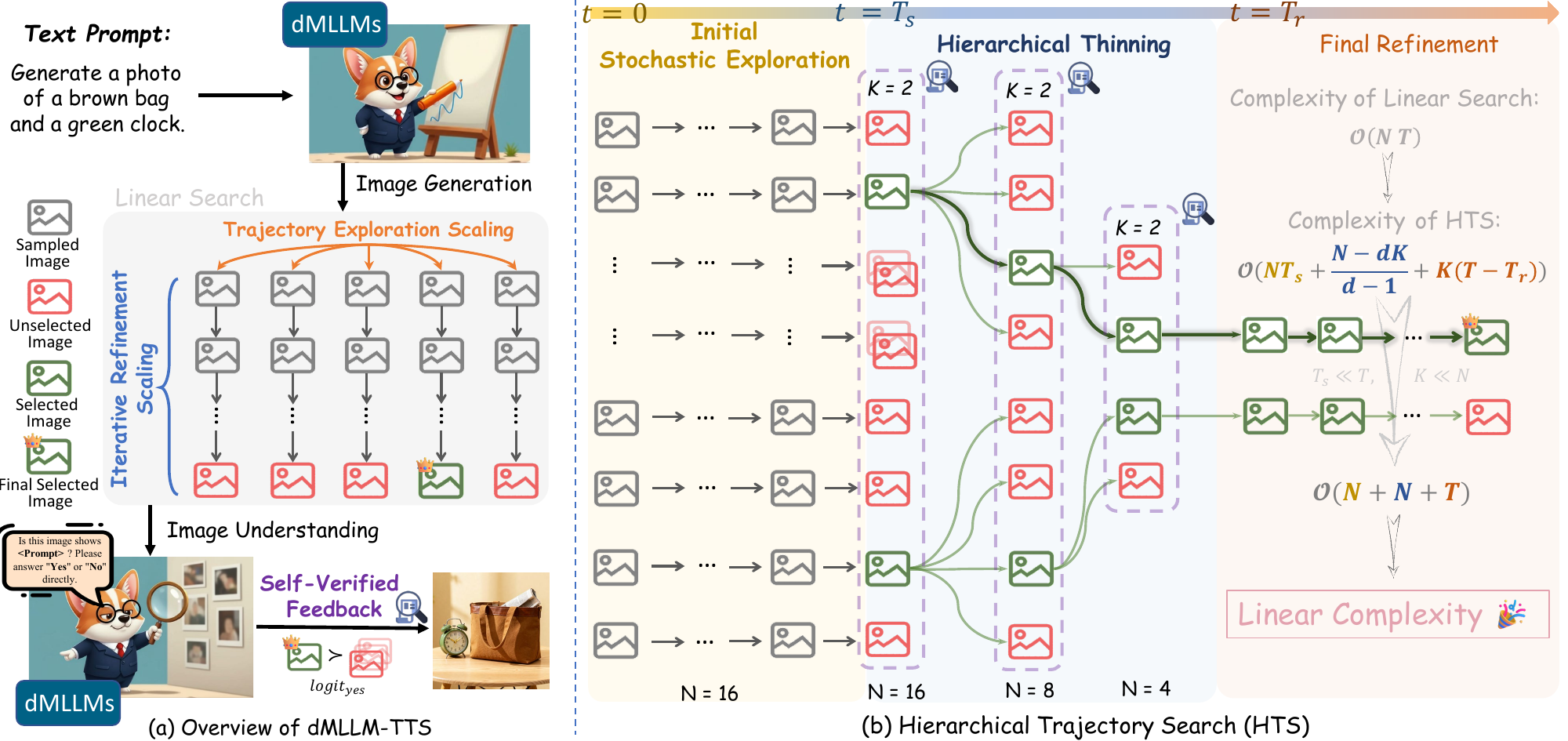}
    \caption{\textbf{Overview of dMLLM-TTS framework.} 
     (a) dMLLM-TTS scales compute along two axes: trajectory exploration and iterative refinement, guided by Self-Verified Feedback for text–image alignment evaluation.
    (b) Hierarchical Trajectory Search (HTS) performs coarse-to-fine generation by starting with broad exploration, pruning low-potential trajectories, and refining high-potential trajectories.}
    \label{fig:overview}
    \vspace{-0.1in}
\end{figure*}

\subsection{How to Scale at Test Time in dMLLMs}
During training, prior text-to-image (T2I) diffusion models learn a denoising vector field that defines a progressive refinement process~\cite{ho2020denoising,rombach2022high}.
At inference time, this mechanism enables scaling through stochastic trajectory sampling and adjustable refinement sampling steps~\cite{ma2025inference}.
dMLLMs inherit this property within a discrete latent token space, where generation proceeds by \textit{iteratively refining token sequences initialized from stochastic states}. 
Thus, test-time scaling in dMLLMs can be viewed as a dual-dimensional process of trajectory exploration and iterative refinement:

\begin{itemize}
    \item \textbf{Trajectory Exploration Scaling.} 
    This dimension enhances the diversity of generated hypotheses. 
    Instead of relying on a single deterministic initialization, the model samples $N$ distinct stochastic latent token sequences $Z_1$ as its starting states (see $t=1$ in Figure~\ref{fig:genprocess}):
    \begin{equation}
        \vspace{-0.05in}
        Z_1 \sim p_{\mathrm{init}}, 
    \end{equation}
    where $p_{\mathrm{init}}$ denotes the initialization prior.
    A larger $N$ broadens the hypothesis space, thereby increasing the likelihood of generating images that are semantically aligned with the given text prompt.
   
    \item \textbf{Iterative Refinement Scaling.} This dimension controls the computational depth per trajectory. 
    Given an initial state $Z_1$, prompt $C$, the model performs $T-1$ refinement steps:
    \begin{equation}
        \vspace{-0.05in}
        Z_{t+1} = \mathcal{G}_{\theta}
        (Z_t, C, t), \quad t = 1,\ldots,T-1.
    \end{equation}
    
    Increasing refinement steps $T$ enhances the granularity of the denoising process, yielding more stable generation trajectories and obtaining higher quality sampled images.

\end{itemize}

While scaling the number of trajectories or refinement steps yields marked improvements in image quality and text-image alignment, it comes at a substantial computational cost.
However, we observe that the potential of a sampling trajectory can be preliminarily assessed during its intermediate process, as illustrated in Figure~\ref{fig:genprocess} (later-stage).
Thus, the key to effective test-time scaling lies not in merely increasing computation,
but in \emph{adaptively allocating it to the most promising trajectories.} 

We formalize this process as an \emph{adaptive trajectory search problem}, 
jointly governed by a generator $\mathcal{G}_{\theta}$, a verifier $\mathcal{V}$, and a search function $f$:
\begin{equation}
\mathrm{TTS} = \langle \mathcal{G}_{\theta}, \mathcal{V}, f \rangle,
\end{equation}

Here, $\mathcal{V}: \mathcal{Z} \times \mathcal{C} \rightarrow \mathbb{R}$ measures semantic alignment between generated images and the given text prompts, and $f$ adaptively redistributes inference computation under $\mathcal{V}$’s guidance. This establish a principled foundation for scalable test-time scaling in dMLLMs.

\subsection{Self-Verified Feedback}
Previous test-time scaling methods often depend on external verifiers, such as CLIP~\cite{radford2021learning}, VILA-Judge~\cite{xie2025sana}, and GPT-4o~\cite{jaech2024openai}, to assess candidate samples.
However, these methods suffer from inherent drawbacks:
\textbf{(i)} they require additional VLM deployments or commercial API calls, increasing resource consumption; and \textbf{(ii)} they incur substantial computational overhead due to repeated decode (converting tokens to image) and encode (transforming images to embeddings for VLMs) operations.

For dMLLMs, the model itself is the ideal verifier $\mathcal{V}$, given its inherent image generation and understanding capabilities.
This lead us to introduce a \textbf{Self-Verified Feedback (SVF)} mechanism, which reuses the model to evaluate generated images via its internal understanding function, enabling efficient, in-loop assessment without external decoding or scoring models, as shown in Figure~\ref{fig:overview}\textcolor{red}{(a)} and following template.

Formally, for a given text prompt~$C$ and an intermediate generated image token sequence $Z_t$,
the model $\mathcal{G}_\theta$ provides binary responses (``yes'' for good quality, ``no'' for low quality) and leverages the logit probability of ``yes'' as the text-image alignment score:
\vspace{-0.05in}
\begin{equation}
\Phi_{\text{SVF}} = \operatorname{logit}_{\text{yes}}\big(\mathcal{G}_\theta(Z_t,C)\big).
\end{equation}
The SVF score thus serves as a semantic measure, allowing the model to rank and select trajectories based on internally consistent criteria.
\begin{tcolorbox}[colback=black!5!white,colframe=black!45!black,title=Instruction Template]
{\textit{\textbf{$<$Generated Image$>$} Is this image shows \textbf{\{text prompt\}? } Please answer ``Yes'' or ``No'' directly without explanation. }}
\end{tcolorbox}

\subsection{How to Efficiently Search the Optimal Sample}
Conventional search strategies, such as the \textbf{Linear Trajectory Search (LTS) }baseline~\cite{xin2025lumina,xie2025sana},
allocate equal inference compute across all trajectories and incur {$\mathcal{O}(NT)$ complexity
without leveraging intermediate sampling feedback, as shown in Figure~\ref{fig:overview}\textcolor{red}{(a)}. 
While effective, this brute-force approach is highly inefficient for dMLLMs' \emph{coarse-to-fine} generative hierarchy, where early stages establish the global structure, while later stages refine semantic details. 
Thus, it is redundant to invest compute resources in trajectories that are already flawed in the early stages.

To address this limitation and achieve
a better balance between image quality and test-time computational efficiency,  we propose \textbf{Hierarchical Trajectory Search (HTS)} strategy with $\mathcal{O}(N+T)$ complexity,
which daptively redistributes computation guided by self-verified feedback (SVF)
and further performs \textit{local neighborhood exploration in the vicinity of high-potential trajectories to guide subsequent refinement}.
Specifically, as shown in Figure~\ref{fig:overview}\textcolor{red}{(b)}, the HTS can be divided into three smoothly evolving stage:
\begin{equation}
HTS \Rightarrow 
\left\{
\begin{alignedat}{2}
&\text{Initial Stochastic Exploration},~~t \le T_s, \\[3pt]
&\text{Hierarchical Thinning},~~~~~~T_s < t \le T_r, \\[3pt]
&\text{Final Refinement}, ~~~~~~~~~~~~~~T_r < t \le T,
\end{alignedat}
\right.
\end{equation}
where $T_s$ and $T_r$ denote the transition steps between the three stages.

\vspace{0.05in}
\noindent {\textbf{Initial Stochastic Exploration.} We begin by sampling $N$ stochastic trajectories from the initialization prior:
\begin{equation}
   Z^{(i)}_1 \sim p_{\text{init}}, \quad i = 1,\dots,N.
\end{equation}
Each trajectory is then denoised over $T_s$ steps to generate a coarse structural hypothesis:
\begin{equation}
    Z^{(i)}_{T_s} = \mathcal{G}_{\theta}^{T_s}\!\big(Z^{(i)}_{1},\, C\big).
\end{equation}
where $\mathcal{G}_\theta^{T_s}$ denotes the $T_s$ steps diffusion propagation operator. 
At this initial, high-noise stage, token representations remain highly stochastic, resulting in blurry generated images (see Figure~\ref{fig:genprocess}). Consequently, SVF-based evaluation is less meaningful at this stage.
Focus should instead be directed towards broad structural exploration and developing a diverse array of initial hypotheses for further refinement.

\vspace{0.05in}
\noindent {\textbf{Progressive Hierarchical Thinning.} We define a progressive \emph{trajectory-width decay schedule} that controls how the search space gradually narrows throughout inference:
\begin{equation}
W_t = \max\!\big(\lfloor N d^{-(t-T_s)}\rfloor,\, K\big), \qquad d > 1,
\end{equation}
where $K$ is the minimal retained set, and $d$ is a decay coefficient determining how aggressively the pool contracts. 
This formulation ensures that the number of active trajectories decreases exponentially until convergence.
At each refinement step $t$, computation is adaptively redistributed toward promising trajectories under the guidance of SVF:

\begin{enumerate}[leftmargin=1.5em]
    \item \textbf{Scoring.}
    Each trajectory in the current pool $\{Z^{(i)}_{t}\}_{i=1}^{W_{t}}$
    is assigned a SVF-based score $\Phi_{\text{SVF}}$.
    
    \item \textbf{Selection.}
    The top-$K$ trajectories with the highest scores form the surviving set $B_t$.

    \item \textbf{Branching.}
    Each survivor $Z^{(j)}_t \in B_t$ then generates $b_t$ stochastic continuations 
    from a local kernel $q$:
    \begin{equation}
    b_t = \Big\lfloor \frac{W_{t+1}}{K}\Big\rfloor, 
    ~~Z^{(j,k)}_{t+1} \sim q\!\big(Z\mid Z^{(j)}_t\big),
    ~~k = 1,\dots,b_t.
    \end{equation}
\end{enumerate}

As $t$ increases, both $W_t$ and $b_t$ decrease geometrically,
incrementally focusing computation on potential trajectories with high SVF scores.
When $W_t = K$ (i.e., $b_t = 1$), branching stops, and the surviving trajectories transition into the final refinement stage,

\definecolor{+}{rgb}{0.70,0.13,0.13}
\definecolor{-}{rgb}{0.25,0.41,0.88}
\definecolor{uclablue}{RGB}{159, 195, 224}
\newcommand{\uclablue}{\cellcolor{uclablue}}

\definecolor{uclagold}{RGB}{254,180,167}
\newcommand{\uclagold}{\cellcolor{uclagold}}

\definecolor{grayred}{RGB}{232,237,205}
\newcommand{\grayred}{\cellcolor{grayred}}

\begin{table*}[!t]
  \centering
  \small
  \renewcommand{\arraystretch}{1.25}
  \setlength{\tabcolsep}{1mm}  
  \vspace{-0.25in}
  \caption{\textbf{Quantitative performance comparison on GenEval} across various d-MLLMs and test-time scaling settings.}
  \vspace{-0.05in}
  \resizebox{\textwidth}{!}{
  \begin{tabular}{c|ccc|cccccc|c}
    \toprule
    \makecell*[l]{} &\textbf{\#Steps} &\textbf{\#Trajectory}  & \textbf{Search} &\makecell*[c]{\textbf{Single Obj.}} & \makecell*[c]{\textbf{Two Obj.}} & \makecell*[c]{\textbf{Counting}} & \makecell*[c]{\textbf{Colors}} & \makecell*[c]{\textbf{Position}} & \makecell*[c]{\textbf{Attribute}} & \textbf{Overall} \\
    \midrule
    \rowcolor{blue!10} 
    \multicolumn{11}{c}{ \textbf{\textit{Text-to-Image Models}}} \\
    \midrule
    \textbf{FLUX.1-dev}~\cite{flux2024} &- &- &- &0.99 & 0.81 &  0.75 &  0.80 &  0.21 &  0.48 & 0.67 \\
    
    \textbf{Janus-Pro}~\cite{chen2025januspro} &- &- &- &$0.99$ &$0.89$ & $0.59$ &$0.90$ &$0.79$ &$0.66$ & $0.80$ \\

    \textbf{BAGEL}~\cite{bagel} &-&-&- &0.99 &0.94 &0.81 &0.88 &0.64 &0.63 &0.82 \\

    \textbf{GPT-4o}~\cite{jaech2024openai} &-&-&- &0.99 &0.92 &0.85 &0.92 &0.75 &0.61 & 0.84 \\

    \textbf{Qwen-Image}~\cite{wu2025qwen} &-&-&- &0.99 &0.92 &0.89 &0.88 &0.76 &0.77 &0.87 \\
    
    \midrule
    \rowcolor{green!10} 
    \multicolumn{11}{c}{ \textbf{\textit{Test-Time Scaling}}} \\
    \midrule
    \rowcolor{gray!12}\textbf{Lumina-DiMOO}~\cite{xin2025luminadimoo} &T = 8 &N = 1 &- 
    &0.96
    &0.85
    &0.70
    &0.85
    &0.72
    &0.62
    &0.78 \\
    
    \cmidrule{1-11}
    \multirow{4}*{\makecell*[l]{\textbf{+ Test-Time Scaling}}}&\multirow{2}*{T = 16}  & N = 16  &LTS 
    &\hspace{0.7cm}{1.0} \textsuperscript{{\scriptsize\color{+}{+4.2\%}}}
    &\hspace{0.75cm}{0.93} \textsuperscript{{\scriptsize\color{+}{+9.4\%}}}
    &\hspace{0.85cm}{0.81} \textsuperscript{{\scriptsize\color{+}{+15.7\%}}}
    &\hspace{0.75cm}{0.90} \textsuperscript{{\scriptsize\color{+}{+5.9\%}}}
    &\hspace{0.85cm}{0.83} \textsuperscript{{\scriptsize\color{+}{+15.3\%}}}
    &\hspace{0.85cm}{0.76} \textsuperscript{{\scriptsize\color{+}{+22.6\%}}} 
    &\hspace{0.85cm}{0.87} \textsuperscript{{\scriptsize\color{+}{+11.5\%}}} \\

    & &(K = 4) &HTS 
    & \hspace{0.7cm}{1.0} \textsuperscript{{\scriptsize\color{+}{+4.2\%}}}
    & \hspace{0.85cm}{0.94} \textsuperscript{{\scriptsize\color{+}{+10.6\%}}}
    & \hspace{0.85cm}{0.84} \textsuperscript{{\scriptsize\color{+}{+20.0\%}}}
    & \hspace{0.75cm}{0.92} \textsuperscript{{\scriptsize\color{+}{+8.2\%}}}
    & \hspace{0.85cm}{0.86} \textsuperscript{{\scriptsize\color{+}{+18.9\%}}}
    & \hspace{0.85cm}{0.79} \textsuperscript{{\scriptsize\color{+}{+27.4\%}}} 
    & \hspace{0.85cm}{0.89} \textsuperscript{{\scriptsize\color{+}{+14.1\%}}} \\

    \cmidrule{2-11} 
    &\multirow{2}*{ T = 32} & N = 32 &LTS
    & \hspace{0.7cm}{1.0} \textsuperscript{{\scriptsize\color{+}{+4.2\%}}} 
    & \hspace{0.85cm}{0.97} \textsuperscript{{\scriptsize\color{+}{+14.1\%}}} 
    & \hspace{0.85cm}{0.89} \textsuperscript{{\scriptsize\color{+}{+27.1\%}}} 
    & \hspace{0.75cm}{0.93} \textsuperscript{{\scriptsize\color{+}{+9.4\%}}} 
    & \hspace{0.9cm}{0.87} \textsuperscript{{\scriptsize\color{+}{+20.8\%}}} 
    & \hspace{0.8cm}{0.80}\textsuperscript{{\scriptsize\color{+}{+29.0\%}}} 
    & \hspace{0.85cm}{0.91} \textsuperscript{{\scriptsize\color{+}{+15.4\%}}} \\

    & & (K = 8) &HTS
    & \hspace{0.7cm}{1.0} \textsuperscript{{\scriptsize\color{+}{+4.2\%}}} 
    & \hspace{0.85cm}{0.97} \textsuperscript{{\scriptsize\color{+}{+14.1\%}}} 
    & \hspace{0.85cm}{0.91} \textsuperscript{{\scriptsize\color{+}{+30.0\%}}} 
    & \hspace{0.85cm}{0.94} \textsuperscript{{\scriptsize\color{+}{+10.6\%}}} 
    & \hspace{0.9cm}{0.89} \textsuperscript{{\scriptsize\color{+}{+23.6\%}}} 
    & \hspace{0.85cm}{0.82} \textsuperscript{{\scriptsize\color{+}{+32.3\%}}} 
    & \hspace{0.85cm}{0.92} \textsuperscript{{\scriptsize\color{+}{+17.9\%}}} \\

    \midrule
    
    \rowcolor{gray!12}\textbf{MMaDA}~\cite{yang2025mmada} &T = 8 &N = 1 &- 
    &0.94 &0.60 &0.32 &0.76 &0.19 &0.22 &0.51 \\
    
    \cmidrule{1-11}
    \multirow{4}*{\makecell*[l]{\textbf{+ Test-Time Scaling}}}&\multirow{2}*{T = 16}  &N = 16  &LTS 
    &\hspace{0.75cm}{0.98} \textsuperscript{{\scriptsize\color{+}{+4.3\%}}} 
    &\hspace{0.85cm}{0.73} \textsuperscript{{\scriptsize\color{+}{+21.7\%}}} 
    &\hspace{0.85cm}{0.47} \textsuperscript{{\scriptsize\color{+}{+46.9\%}}} 
    &\hspace{0.85cm}{0.88} \textsuperscript{{\scriptsize\color{+}{+15.8\%}}} 
    &\hspace{0.85cm}{0.25} \textsuperscript{{\scriptsize\color{+}{+31.6\%}}} 
    &\hspace{0.85cm}{0.31} \textsuperscript{{\scriptsize\color{+}{+42.9\%}}} 
    &\hspace{0.85cm}{0.60} \textsuperscript{{\scriptsize\color{+}{+17.6\%}}} \\ 

    & & (K = 4) &HTS 
    &\hspace{0.75cm}{0.99} \textsuperscript{{\scriptsize\color{+}{+5.3\%}}}
    &\hspace{0.85cm}{0.75} \textsuperscript{{\scriptsize\color{+}{+25.0\%}}}
    &\hspace{0.85cm}{0.48} \textsuperscript{{\scriptsize\color{+}{+50.0\%}}}
    &\hspace{0.85cm}{0.90} \textsuperscript{{\scriptsize\color{+}{+18.4\%}}}
    &\hspace{0.85cm}{0.27} \textsuperscript{{\scriptsize\color{+}{+42.1\%}}}
    &\hspace{0.85cm}{0.32} \textsuperscript{{\scriptsize\color{+}{+45.5\%}}}
    &\hspace{0.85cm}{0.62} \textsuperscript{{\scriptsize\color{+}{+21.6\%}}}\\

    \cmidrule{2-11}
    &\multirow{2}*{T = 32} & N = 32 &LTS
    & \hspace{0.75cm}{0.99} \textsuperscript{{\scriptsize\color{+}{+5.3\%}}}
    & \hspace{0.85cm}{0.77} \textsuperscript{{\scriptsize\color{+}{+28.3\%}}}
    & \hspace{0.85cm}{0.50} \textsuperscript{{\scriptsize\color{+}{+56.2\%}}}
    & \hspace{0.85cm}{0.91} \textsuperscript{{\scriptsize\color{+}{+19.7\%}}}
    & \hspace{0.85cm}{0.29} \textsuperscript{{\scriptsize\color{+}{+52.6\%}}}
    & \hspace{0.85cm}{0.36} \textsuperscript{{\scriptsize\color{+}{+63.6\%}}}
    & \hspace{0.85cm}{0.64} \textsuperscript{{\scriptsize\color{+}{+25.5\%}}}\\

    & &(K = 8) &HTS
    & \hspace{0.75cm}{0.99} \textsuperscript{{\scriptsize\color{+}{+5.3\%}}}
    & \hspace{0.85cm}{0.81} \textsuperscript{{\scriptsize\color{+}{+35.0\%}}}
    & \hspace{0.85cm}{0.52} \textsuperscript{{\scriptsize\color{+}{+62.5\%}}}
    & \hspace{0.85cm}{0.92} \textsuperscript{{\scriptsize\color{+}{+21.1\%}}}
    & \hspace{0.85cm}{0.33} \textsuperscript{{\scriptsize\color{+}{+73.6\%}}}
    & \hspace{0.85cm}{0.37} \textsuperscript{{\scriptsize\color{+}{+68.2\%}}}
    & \hspace{0.85cm}{0.66} \textsuperscript{{\scriptsize\color{+}{+29.4\%}}}
    \\
    \midrule

    \rowcolor{gray!12}\textbf{Muddit}~\cite{shi2025muddit} &T = 8 &N = 1 &- 
    & 0.95 &0.53 &0.48 &0.80 &0.13 &0.29 &0.53 \\
    
    \cmidrule{1-11}
    \multirow{4}*{\makecell*[l]{\textbf{+ Test-Time Scaling}}}&\multirow{2}*{T = 16}  & N = 16  &LTS 
    & \hspace{0.75cm}{0.98}  \textsuperscript{{\scriptsize\color{+}{+3.2\%}}}
    & \hspace{0.85cm}{0.68} \textsuperscript{{\scriptsize\color{+}{+28.3\%}}}
    & \hspace{0.85cm}{0.59} \textsuperscript{{\scriptsize\color{+}{+22.9\%}}}
    & \hspace{0.8cm}{0.86} \textsuperscript{{\scriptsize\color{+}{+7.5\%}}}
    & \hspace{0.85cm}{0.22} \textsuperscript{{\scriptsize\color{+}{+69.2\%}}}
    & \hspace{0.85cm}{0.39} \textsuperscript{{\scriptsize\color{+}{+34.5\%}}}
    & \hspace{0.85cm}{0.62} \textsuperscript{{\scriptsize\color{+}{+17.0\%}}}
    \\

    & &(K = 4) &HTS
    & \hspace{0.75cm}{0.99}  \textsuperscript{{\scriptsize\color{+}{+4.2\%}}}
    & \hspace{0.85cm}{0.71} \textsuperscript{{\scriptsize\color{+}{+33.9\%}}}
    & \hspace{0.85cm}{0.59} \textsuperscript{{\scriptsize\color{+}{+22.9\%}}}
    & \hspace{0.85cm}{0.88} \textsuperscript{{\scriptsize\color{+}{+10.0\%}}}
    & \hspace{0.85cm}{0.24} \textsuperscript{{\scriptsize\color{+}{+84.6\%}}}
    & \hspace{0.85cm}{0.41} \textsuperscript{{\scriptsize\color{+}{+41.4\%}}}
    & \hspace{0.85cm}{0.64} \textsuperscript{{\scriptsize\color{+}{+20.8\%}}}
    \\
    \cmidrule{2-11}
    &\multirow{2}*{T = 32} &N = 32 &LTS
    & \hspace{0.75cm}{0.99}  \textsuperscript{{\scriptsize\color{+}{+4.2\%}}}
    & \hspace{0.85cm}{0.73} \textsuperscript{{\scriptsize\color{+}{+37.3\%}}}
    & \hspace{0.85cm}{0.60} \textsuperscript{{\scriptsize\color{+}{+25.0\%}}}
    & \hspace{0.85cm}{0.89} \textsuperscript{{\scriptsize\color{+}{+11.2\%}}}
    & \hspace{0.85cm}{0.24} \textsuperscript{{\scriptsize\color{+}{+84.6\%}}}
    & \hspace{0.85cm}{0.42} \textsuperscript{{\scriptsize\color{+}{+44.8\%}}}
    & \hspace{0.85cm}{0.65} \textsuperscript{{\scriptsize\color{+}{+22.6\%}}}
    \\

    & & (K = 8) &HTS
    & \hspace{0.75cm}{0.99}  \textsuperscript{{\scriptsize\color{+}{+4.2\%}}}
    & \hspace{0.85cm}{0.76} \textsuperscript{{\scriptsize\color{+}{+43.4\%}}}
    & \hspace{0.85cm}{0.62} \textsuperscript{{\scriptsize\color{+}{+29.2\%}}}
    & \hspace{0.85cm}{0.88} \textsuperscript{{\scriptsize\color{+}{+10.0\%}}}
    & \hspace{0.85cm}{0.25} \textsuperscript{{\scriptsize\color{+}{+92.3\%}}}
    & \hspace{0.85cm}{0.44} \textsuperscript{{\scriptsize\color{+}{+51.7\%}}}
    & \hspace{0.85cm}{0.67} \textsuperscript{{\scriptsize\color{+}{+26.4\%}}}
    \\
    \bottomrule
  \end{tabular}
  }
  \label{tab:main_tts_results}
\end{table*}
\begin{figure*}[!t]
    \centering
    \includegraphics[width=1.0\linewidth]{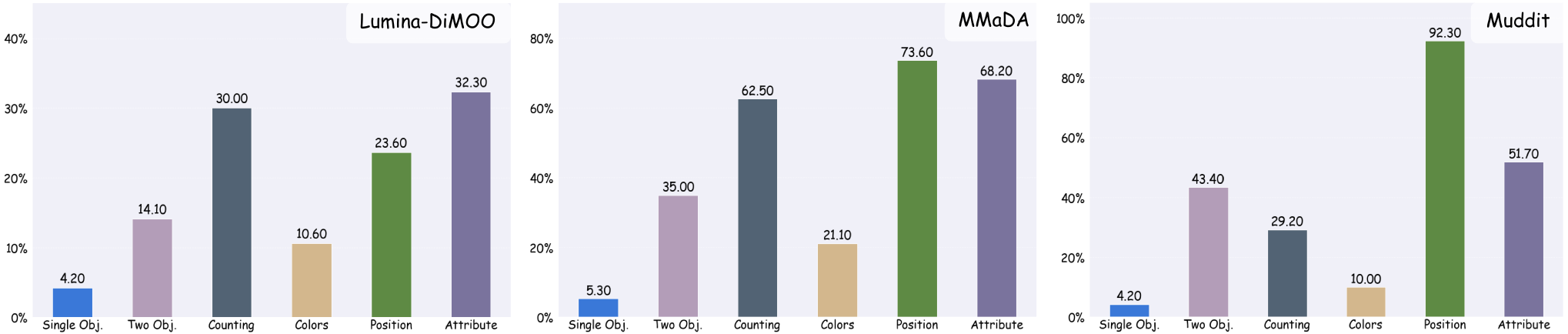}
    \vspace{-0.17in}
    \caption{\textbf{Qualitative improvement ratio in TTS performance across various text prompt complexities} examined through diverse dMLLMs on GenEval benchmark dimensions. TTS markedly enhances performance across all measured dimensions.}
    \label{fig:ab_dimension}
    \vspace{-0.05in}
\end{figure*}

\vspace{0.05in}
\noindent {\textbf{Final Refinement.} 
The adaptive thinning process contracts the trajectory pool to a minimum width of $K$ at step $T_r$, signifying the transition from trajectory exploration to refinement.
Subsequently, the $K$ surviving trajectories are refined independently until the final diffusion step $T$:
\begin{equation}
Z_{T}^{(j)} = \mathcal{G}_{\theta}^{\,T - T_r}\!\big(Z_{T_r}^{(j)},\, C\big),
\qquad j = 1,\dots,K.
\end{equation}
At this stage, all computational resources are redirected to enhancing the image’s details.
By concentrating its inference depth on the text-image aligned trajectories identified by SVF, the model transforms early stochastic exploration into precise, text-aligned outputs.
This adaptive contraction strategy dynamically selects the most promising paths for detailed refinement, effectively merging the benefits of broad exploration and deep refinement within a single, compute-efficient inference process.

\vspace{0.05in}
\noindent {\textbf{Complexity Analysis.}
The total forward cost of HTS can be expressed as:
\begin{equation}
C_{\text{HTS}} =
\mathcal{O}(N T_s
+ \frac{N - dK}{d - 1}
+ K (T - T_r)),
\end{equation}
where the three terms respectively correspond to 
(i) early stochastic exploration over $N$ trajectories for $T_s$ steps, 
(ii) hierarchical thinning with geometric decay factor $d>1$, 
and (iii) final refinement over the $K$ surviving trajectories.

In practice, we start from a wide exploration with a large number of trajectories $N$ 
and quickly prune to a compact set ($K\!\ll\!N$), while the early warm-up stage 
($T_s\!\ll\!T$) only covers a small portion of the diffusion process to establish structural diversity. 
Hence the complexity simplifies to:
\begin{equation}
C_{\text{HTS}}
\;\approx\;
\mathcal{O}(N + T).
\end{equation}

This near-linear scaling demonstrates that HTS effectively reallocates computation 
from early wide exploration to late fine refinement, 
avoiding the quadratic $\mathcal{O}(NT)$ cost of linear trajectory search (LTS). 
By performing coarse structural exploration under high noise and 
gradually contracting the trajectory pool to $K$ candidates for fine-grained synthesis, 
HTS efficiently balances search width and depth during inference.

\begin{figure*}[!t]
    \centering
    \vspace{-0.1in}
    \includegraphics[width=1.0\linewidth]{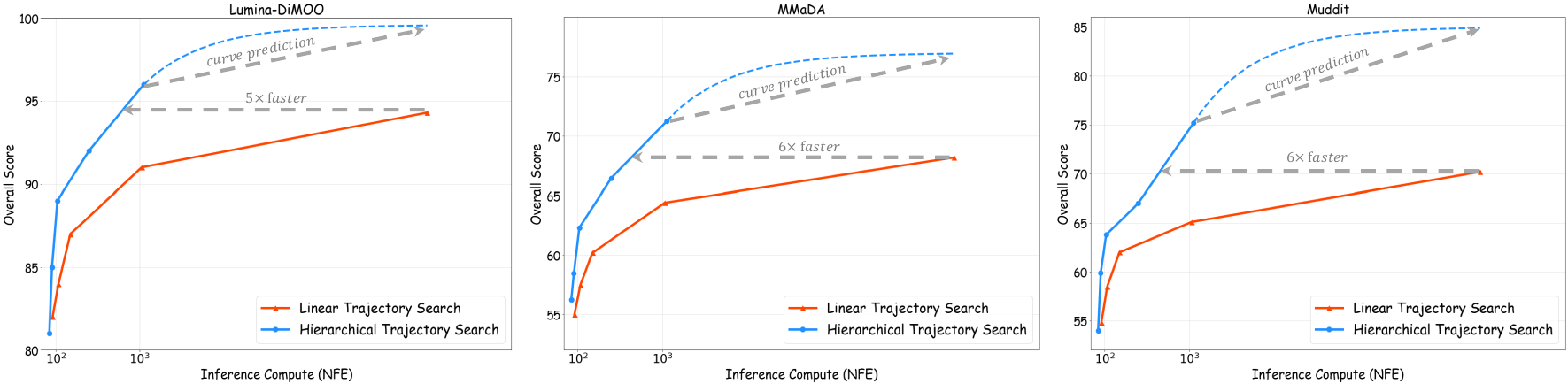}
    \vspace{-0.2in}
    \caption{\textbf{Comparison between linear and hierarchical trajectory search.} The red curve illustrates linear trajectory search, while the blue curve depicts hierarchical trajectory search, with a dashed line indicating predictions based on a geometric series decay approximation. 
    Curve fitting shows that similar subsequent trends tend to converge towards an upper limit.}
    \label{fig:nfe}
    \vspace{-0.1in}
\end{figure*}
\section{Experiment}
\label{sec:exp}
\subsection{Experiment Setup}
\paragraph{Evaluation Models.}
We validate our proposed test-time scaling method (\textit{i.e.}, scaling strategy, verifier, and search algorithm) using three popular open-source pre-trained dMLLMs, including Lumina-DiMOO~\cite{xin2025luminadimoo}, MMaDA~\cite{yang2025mmada}, and Muddit~\cite{shi2025muddit}. 
These models span a parameter range from 1B to 8B, which allows us to verify the generality of our method across different model scales. Due to the lack of open-source access to other notable dMLLMs, such as Lavida-O, it is currently not feasible to conduct experiments with them.

\vspace{0.05in}
\noindent {\textbf{Evaluation Metrics.}
To quantify the performance of text-to-image (T2I) generation, we employ the GenEval~\cite{ghosh2024geneval} benchmark, which is specifically designed to assess object-centric generation using compositional prompts with varied object attributes. 
GenEval, consisting of 553 text prompts, is the most widely used benchmark within the TTS domain.  
To measure computational cost, we adopt the metric of the number of function evaluations (NFE), consistent with previous TTS studies on generative models~\cite{ma2025inference,xin2025lumina,zhuo2025reflection}.

\vspace{0.05in}
\noindent {\textbf{Implementation Details.}
The baseline performance for each model is established using 8 sampling steps.
To ensure a fair comparison, all images are generated at a 512$\times$512 resolution with CFG=4.0.
For image verification, the model performs one step to produce a ``Yes" or ``No" response. 
In addition to our proposed Self-Verified Feedback, we conducted further experiments using external verifiers, as detailed in Section~\ref{para:verifier}.
For the HTS strategy, the primary experimental setting maintains an N:K ratio of 4:1.
As for $T_s$ in the HTS, it is empirically set to T/4, while $T_r$ is determined by N and K.

\subsection{Main Results}
\paragraph{TTS Consistently Delivers Stable Performance Gains Across Various d-MLLMs.} 
We applied our dMLLM-TTS method to the Lumina-DiMOO, MMaDA, and Muddit models, yielding significant and consistent performance gains on the GenEval benchmark (Table~\ref{tab:main_tts_results}).
Specifically, Lumina-DiMOO’s score rose from 0.78 to 0.92 (+17.9\%), MMaDA’s from 0.51 to 0.66 (+29.4\%), and Muddit’s from 0.53 to 0.67 (+26.4\%).
This enhancement elevates their capabilities to match or even surpass SOTA text-to-image models.
For example, the TTS-enhanced Lumina-DiMOO (0.92) outperforms leading models such as Qwen-Image (0.87) and GPT-4o (0.84).
These findings demonstrate that dMLLM-TTS is an effective and widely applicable method for improving the generative quality of d-MLLMs.

\vspace{0.05in}
\noindent {\textbf{TTS Significantly Boosts Performance Across All Dimensions.} 
The ability to handle text prompts of varying complexity across multiple dimensions is a key challenge of T2I generation.
Our dMLLM-TTS consistently boosts performance across six dimensions, as shown in Figure~\ref{fig:ab_dimension}.
For common prompt sets (\textit{e.g.}, Single Object, Two Object, Colors), dMLLM-TTS provides notable improvements, although the gains are inherently limited by the strong baseline performance.
Crucially, for complex prompt sets (\textit{e.g.}, Counting, Position, Attribute), dMLLM-TTS achieves substantial performance gains.
This highlights its effectiveness in elevating the dMLLMs’ performance ceiling, especially on challenging compositional generation tasks.

\vspace{0.05in}
\noindent {\textbf{HTS Proves More Efficient and Superior to LTS Baseline.} 
We evaluated our Hierarchical Trajectory Search (HTS) against the Linear Trajectory Search (LTS) baseline in Figure~\ref{fig:nfe}. 
The results reveal clear advantages for our HTS approach in both efficiency and performance.
HTS requires substantially less inference compute to reach equivalent scores, achieving 5$\times$ faster on Lumina-DiMOO and 6$\times$ faster on MMaDA and Muddit. 
Furthermore, beyond just accelerating the search, HTS consistently converges to a higher overall score than the baseline.
These findings confirm that HTS offers a powerful combination of computational savings and superior results.

\subsection{Analysis and Discussion}
\begin{figure*}[!t]
    \centering
    \vspace{-0.05in}
    \includegraphics[width=1.0\linewidth]{Figure/ab1.pdf}
    \vspace{-0.15in}
    \caption{\textbf{Trajectory Exploration Scaling (Left) and Iterative Refinement Scaling (Right).}
    Increasing the number of explored trajectories (N = 1→32) refinement steps (T = 8→64) consistently improves performance across all dMLLMs.}
    \label{fig:abation_trajectory_steps}
\end{figure*}

\begin{figure*}[!t]
    \centering
    \vspace{-0.02in}
    \includegraphics[width=1.01\linewidth]{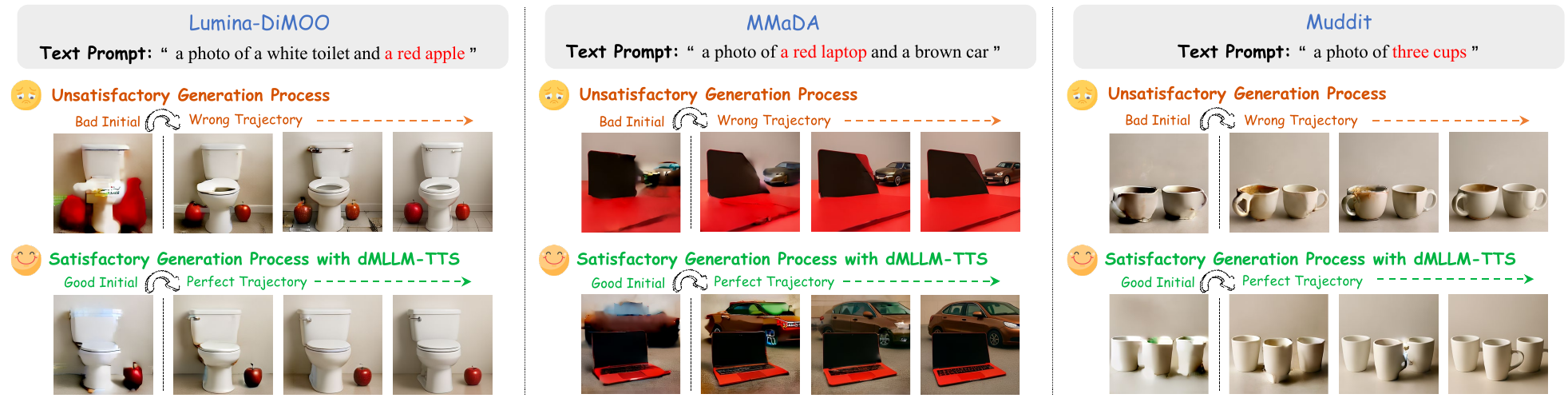}
    \vspace{-0.15in}
    \caption{\textbf{Image Generation Process without (Top) and with (Bottom) dMLLM-TTS.} The baseline models produce unsatisfactory text-to-image results. However, by incorporating our TTS strategies, the generation process is significantly improved.}
    \label{fig:case}
    \vspace{-0.1in}
\end{figure*}

\paragraph{Effect of Trajectory Exploration Scaling.} 
As depicted in Figure~\ref{fig:abation_trajectory_steps} (left), our analysis validates the effectiveness of Trajectory Exploration Scaling for improving GenEval scores. 
With a fixed step of 32, we systematically scaled the exploration trajectory from $N = 1$ to $N = 32$, observing a clear, positive impact across all evaluated models.
Notably, the performance gains are inversely correlated with the models’ initial scores. The baseline models, MMaDA and Muddit, achieved the most significant improvements of +20.2\% and +16.8\%, respectively. Even the top-performing Lumina-DiMOO model benefited from a substantial +8.8\% score increase.
This evidence strongly suggests that our trajectory exploration scaling method is a robust and valuable technique for enhancing model performance

\vspace{0.05in}
\noindent {\textbf{Effect of Iterative Refinement Scaling.} Figure~\ref{fig:abation_trajectory_steps} (right) demonstrates the effect of iterative refinement scaling across three models. 
As the number of refinement steps increases from 8 to 64, overall scores consistently improve, underscoring the effectiveness of our iterative refinement scaling strategy.
However, indefinite scaling is not practical. The ideal number of refinement steps depends on the complexity of the text prompt.
The optimal number of refinement steps typically depends on the complexity of the text prompt. 
For dMLLMs sampling at a resolution of 512$\times$512, up to 64 steps typically provides a strong balance between performance gains and computational efficiency.

\begin{table}[!t]
  \caption{\textbf{Comparison results of various verifiers.}}
  \vspace{-3pt}
  \footnotesize
  \centering
    \begin{tabular}{c|>{\centering\arraybackslash}m{2.2cm}|>{\centering\arraybackslash}m{1.35cm}|>{\centering\arraybackslash}m{1.35cm}}
      \toprule
      \textbf{Verifier} & \textbf{Lumina-DiMOO} & \textbf{MMaDA} & \textbf{Muddit}\\
      \midrule
      \textbf{SVF (Ours)} &0.92 &0.66 & 0.67 \\  
      \textbf{VILA-Judge} &~~~0.90 \textcolor{green}{$\downarrow$} &~~~0.70 \textcolor{red}{$\uparrow$}  &~~~0.70 \textcolor{red}{$\uparrow$}\\
      \textbf{GPT-4o} &~~~0.95 \textcolor{red}{$\uparrow$} &~~~0.71 \textcolor{red}{$\uparrow$} &~~~0.74 \textcolor{red}{$\uparrow$}  \\
      \bottomrule
    \end{tabular}
    \label{tab:verifier}
  \vspace{-1em}
\end{table}
\vspace{0.05in}
\noindent {\textbf{Comparison with External Verifiers.}
\label{para:verifier}
Our analysis shows that external verifiers, particularly powerful commercial model GPT-4o, outperform our self-verified feedback approach, as reported in Table~\ref{tab:verifier}. 
This gap is primarily attributable to their superior image understanding capabilities. 
These results suggest that current dMLLMs still have considerable room to improve in visual comprehension.

\vspace{0.05in}
\noindent {\textbf{Qualitative Examples.} To intuitively illustrate the effectiveness of dMLLM-TTS, we present qualitative examples of generated images and corresponding generation processes. As shown in Figure~\ref{fig:case}, baseline models often produce a bad initial state and follow a wrong generation trajectory. In contrast, our dMLLM-TTS leads to a good initial state and maintain a perfect trajectory. This underscores the significant improvements achieved with dMLLM-TTS, resulting in more precise and faithful image generations that align with the text prompt.

\section{Conclusion}
In this paper, we present a novel TTS framework specifically designed for dMLLMs. By integrating Self-Verified Feedback for internal evaluation and Hierarchical Trajectory Search for adaptive inference, our framework eliminates reliance on external verifiers and optimizes computational allocation. Experiments show that dMLLM-TTS framework significantly enhances text-image alignment, especially for complex prompts, establishing an efficient and powerful paradigm for scalable inference within dMLLMs.

{
    \small
    \bibliographystyle{ieeenat_fullname}
    \bibliography{main}

@String(CVPR= {IEEE Conf. Comput. Vis. Pattern Recog.})

@String(ICCV= {Int. Conf. Comput. Vis.})

@String(NIPS= {Adv. Neural Inform. Process. Syst.})

@String(ICLR = {Int. Conf. Learn. Represent.})

@String{cvpr = "Proceedings of the IEEE Conference on Computer Vision and Pattern Recognition (CVPR)"}

@String{nips = "Advances in Neural Information Processing Systems (NeurIPS)"}

@String{iclr = "Proceedings of the International Conference on Learning Representations (ICLR)"}

@String{iccv = "Proceedings of the IEEE International Conference on Computer Vision (ICCV)"}

@String{icml = "Proceedings of the International Conference on Machine Learning (ICML)"}

@String{emnlp = "Proceedings of the Conference on Empirical Methods in Natural Language Processing (EMNLP)"}

@article{yi2024towards,
  title={Towards Understanding the Working Mechanism of Text-to-Image Diffusion Model},
  author={Yi, Mingyang and Li, Aoxue and Xin, Yi and Li, Zhenguo},
  journal=nips,
  year={2024}
}

@article{ho2020denoising,
  title={Denoising diffusion probabilistic models},
  author={Ho, Jonathan and Jain, Ajay and Abbeel, Pieter},
  journal=nips,
  year={2020}
}

@misc{flux2024,
    author={Black Forest Labs},
    title={FLUX},
    year={2024},
    howpublished={\url{https://github.com/black-forest-labs/flux}},
}

@inproceedings{gaolumina,
  title={Lumina-T2X: Scalable Flow-based Large Diffusion Transformer for Flexible Resolution Generation},
  author={Gao, Peng and Zhuo, Le and Liu, Dongyang and Du, Ruoyi and Luo, Xu and Qiu, Longtian and Zhang, Yuhang and Huang, Rongjie and Geng, Shijie and Zhang, Renrui and others},
  booktitle=iclr,
  year={2025}
}

@article{chen2023pixart,
  title={PixArt-$\Alpha$: Fast Training of Diffusion Transformer for Photorealistic Text-to-Image Synthesis},
  author={Chen, Junsong and Yu, Jincheng and Ge, Chongjian and Yao, Lewei and Xie, Enze and Wu, Yue and Wang, Zhongdao and Kwok, James and Luo, Ping and Lu, Huchuan and others},
  journal=iclr,
  year={2023}
}

@article{patil2024amused,
  title={amused: An open muse reproduction},
  author={Patil, Suraj and Berman, William and Rombach, Robin and von Platen, Patrick},
  journal={arXiv preprint arXiv:2401.01808},
  year={2024}
}

@article{nie2025llada,
  title={Large Language Diffusion Models}, 
  author={Shen Nie and Fengqi Zhu and Zebin You and Xiaolu Zhang and Jingyang Ou and Jun Hu and Jun Zhou and Yankai Lin and Ji-Rong Wen and Chongxuan Li},
  journal={arXiv preprint arXiv:2502.09992},
  year={2025}
}

@article{wang2024emu3,
  title={Emu3: Next-Token Prediction is All You Need},
  author={Wang, Xinlong and Zhang, Xiaosong and Luo, Zhengxiong and Sun, Quan and Cui, Yufeng and Wang, Jinsheng and Zhang, Fan and Wang, Yueze and Li, Zhen and Yu, Qiying and others},
  journal={arXiv preprint arXiv:2409.18869},
  year={2024}
}

@inproceedings{zhuolumina,
  title={Lumina-Next: Making Lumina-T2X Stronger and Faster with Next-DiT},
  author={Zhuo, Le and Du, Ruoyi and Xiao, Han and Li, Yangguang and Liu, Dongyang and Huang, Rongjie and Liu, Wenze and Zhu, Xiangyang and Wang, Fu-Yun and Ma, Zhanyu and others},
  booktitle=nips,
  year={2024}
}

@article{xie2024sana,
  title={Sana: Efficient high-resolution image synthesis with linear diffusion transformers},
  author={Xie, Enze and Chen, Junsong and Chen, Junyu and Cai, Han and Tang, Haotian and Lin, Yujun and Zhang, Zhekai and others},
  journal=iclr,
  year={2025}
}

@article{xie2025sana,
  title={SANA 1.5: Efficient Scaling of Training-Time and Inference-Time Compute in Linear Diffusion Transformer},
  author={Xie, Enze and Chen, Junsong and others},
  journal=icml,
  year={2025}
}

@article{ma2025inference,
  title={Inference-time scaling for diffusion models beyond scaling denoising steps},
  author={Ma, Nanye and Tong, Shangyuan and Jia, Haolin and Hu, Hexiang and Su, Yu-Chuan and Zhang, Mingda and Yang, Xuan and Li, Yandong and Jaakkola, Tommi and Jia, Xuhui and others},
  journal={arXiv preprint arXiv:2501.09732},
  year={2025}
}

@article{ghosh2024geneval,
  title={Geneval: An object-focused framework for evaluating text-to-image alignment},
  author={Ghosh, Dhruba and Hajishirzi, Hannaneh and Schmidt, Ludwig},
  journal=nips,
  volume={36},
  year={2024}
}

@article{liu2024lumina,
  title={Lumina-mgpt: Illuminate flexible photorealistic text-to-image generation with multimodal generative pretraining},
  author={Liu, Dongyang and Zhao, Shitian and Zhuo, Le and Lin, Weifeng and Qiao, Yu and Li, Hongsheng and Gao, Peng},
  journal={arXiv preprint arXiv:2408.02657},
  year={2024}
}

@article{jaech2024openai,
  title={Openai o1 system card},
  author={Jaech, Aaron and Kalai, Adam and Lerer, Adam and Richardson, Adam and El-Kishky, Ahmed and Low, Aiden and Helyar, Alec and Madry, Aleksander and Beutel, Alex and Carney, Alex and others},
  journal={arXiv preprint arXiv:2412.16720},
  year={2024}
}

@article{podell2023sdxl,
  title={Sdxl: Improving latent diffusion models for high-resolution image synthesis},
  author={Podell, Dustin and English, Zion and Lacey, Kyle and Blattmann, Andreas and Dockhorn, Tim and M{\"u}ller, Jonas and Penna, Joe and Rombach, Robin},
  journal=iclr,
  year={2023}
}

@article{xie2024show,
  title={Show-o: One single transformer to unify multimodal understanding and generation},
  author={Xie, Jinheng and Mao, Weijia and Bai, Zechen and Zhang, David Junhao and Wang, Weihao and Lin, Kevin Qinghong and Gu, Yuchao and Chen, Zhijie and Yang, Zhenheng and Shou, Mike Zheng},
  journal={arXiv preprint arXiv:2408.12528},
  year={2024}
}

@article{chen2025januspro,
  title={Janus-Pro: Unified Multimodal Understanding and Generation with Data and Model Scaling}, 
  author={Xiaokang Chen and Zhiyu Wu and Xingchao Liu and Zizheng Pan and Wen Liu and Zhenda Xie and Xingkai Yu and Chong Ruan},
  journal={arXiv preprint arXiv:2501.17811},
  year={2025},
}

@inproceedings{rombach2022high,
  title={High-resolution image synthesis with latent diffusion models},
  author={Rombach, Robin and Blattmann, Andreas and Lorenz, Dominik and Esser, Patrick and Ommer, Bj{\"o}rn},
  booktitle=cvpr,
  year={2022}
}

@inproceedings{hessel2021clipscore,
  title={Clipscore: A reference-free evaluation metric for image captioning},
  author={Hessel, Jack and Holtzman, Ari and Forbes, Maxwell and Le Bras, Ronan and Choi, Yejin},
  booktitle=emnlp,
  year={2021}
}

@article{qin2025lumina,
  title={Lumina-Image 2.0: A Unified and Efficient Image Generative Framework},
  author={Qin, Qi and Zhuo, Le and Xin, Yi and Du, Ruoyi and Li, Zhen and Fu, Bin and Lu, Yiting and Yuan, Jiakang and Li, Xinyue and Liu, Dongyang and others},
  journal=iccv,
  year={2025}
}

@article{hurst2024gpt,
  title={Gpt-4o system card},
  author={Hurst, Aaron and Lerer, Adam and Goucher, Adam P and Perelman, Adam and Ramesh, Aditya and Clark, Aidan and Ostrow, AJ and Welihinda, Akila and Hayes, Alan and Radford, Alec and others},
  journal={arXiv preprint arXiv:2410.21276},
  year={2024}
}

@article{li2025reflect,
  title={Reflect-DiT: Inference-Time Scaling for Text-to-Image Diffusion Transformers via In-Context Reflection},
  author={Li, Shufan and Kallidromitis, Konstantinos and Gokul, Akash and Koneru, Arsh and Kato, Yusuke and Kozuka, Kazuki and Grover, Aditya},
  journal={arXiv preprint arXiv:2503.12271},
  year={2025}
}

@article{zhuo2025reflection,
  title={From reflection to perfection: Scaling inference-time optimization for text-to-image diffusion models via reflection tuning},
  author={Zhuo, Le and Zhao, Liangbing and Paul, Sayak and Liao, Yue and Zhang, Renrui and Xin, Yi and others},
  journal=iccv,
  year={2025}
}

@article{xin2025luminadimoo,
  title={Lumina-dimoo: An omni diffusion large language model for multi-modal generation and understanding},
  author={Xin, Yi and Qin, Qi and Luo, Siqi and Zhu, Kaiwen and Yan, Juncheng and Tai, Yan and Lei, Jiayi and Cao, Yuewen and Wang, Keqi and Wang, Yibin and others},
  journal={arXiv preprint arXiv:2510.06308},
  year={2025}
}

@article{xin2025lumina,
  title={Lumina-mGPT 2.0: Stand-Alone AutoRegressive Image Modeling},
  author={Xin, Yi and Yan, Juncheng and Qin, Qi and Li, Zhen and Liu, Dongyang and Li, Shicheng and Huang, Victor Shea-Jay and Zhou, Yupeng and Zhang, Renrui and Zhuo, Le and others},
  journal={arXiv preprint arXiv:2507.17801},
  year={2025}
}

@article{yang2025mmada,
  title={Mmada: Multimodal large diffusion language models},
  author={Yang, Ling and Tian, Ye and Li, Bowen and Zhang, Xinchen and Shen, Ke and Tong, Yunhai and Wang, Mengdi},
  journal={arXiv preprint arXiv:2505.15809},
  year={2025}
}

@article{xin2025resurrect,
  title={Resurrect mask autoregressive modeling for efficient and scalable image generation},
  author={Xin, Yi and Zhuo, Le and Qin, Qi and Luo, Siqi and Cao, Yuewen and Fu, Bin and He, Yangfan and Li, Hongsheng and Zhai, Guangtao and Liu, Xiaohong and others},
  journal={arXiv preprint arXiv:2507.13032},
  year={2025}
}

@article{singhal2025general,
  title={A general framework for inference-time scaling and steering of diffusion models},
  author={Singhal, Raghav and Horvitz, Zachary and Teehan, Ryan and Ren, Mengye and Yu, Zhou and others},
  journal={arXiv preprint arXiv:2501.06848},
  year={2025}
}

@article{chen2025tts,
  title={TTS-VAR: A Test-Time Scaling Framework for Visual Auto-Regressive Generation},
  author={Chen, Zhekai and Chu, Ruihang and Chen, Yukang and Zhang, Shiwei and Wei, Yujie and Zhang, Yingya and Liu, Xihui},
  journal={arXiv preprint arXiv:2507.18537},
  year={2025}
}

@inproceedings{radford2021learning,
  title={Learning transferable visual models from natural language supervision},
  author={Radford, Alec and Kim, Jong Wook and Hallacy, Chris and Ramesh, Aditya and Goh, Gabriel and Agarwal, Sandhini and Sastry, Girish and Askell, Amanda and Mishkin, Pamela and Clark, Jack and others},
  booktitle=icml,
  year={2021}
}

@inproceedings{austin2021d3pm,
  title     = {Structured Denoising Diffusion Models in Discrete State-Spaces},
  author    = {Jacob Austin and Daniel D. Johnson and Jonathan Ho and Daniel Tarlow and Rianne van den Berg},
  booktitle = {Advances in Neural Information Processing Systems (NeurIPS)},
  year      = {2021}
}

@inproceedings{lou2023sedd,
  title={Discrete diffusion modeling by estimating the ratios of the data distribution},
  author={Lou, Aaron and Meng, Chenlin and Ermon, Stefano},
  booktitle=icml,
  year={2024}
}

@inproceedings{gong2024scalingdlm,
  title={Scaling Diffusion Language Models via Adaptation from Autoregressive Models},
  author={Gong, Shansan and Agarwal, Shivam and Zhang, Yizhe and Ye, Jiacheng and Zheng, Lin and Li, Mukai and An, Chenxin and Zhao, Peilin and Bi, Wei and Han, Jiawei and others},
  booktitle=iclr,
  year= {2025}
}

@misc{shi2025muddit,
      title={Muddit: Liberating Generation Beyond Text-to-Image with a Unified Discrete Diffusion Model}, 
      author={Qingyu Shi and Jinbin Bai and Zhuoran Zhao and Wenhao Chai and Kaidong Yu and Jianzong Wu and Shuangyong Song and Yunhai Tong and Xiangtai Li and Xuelong Li and Shuicheng Yan},
      year={2025},
      eprint={2505.23606},
      archivePrefix={arXiv},
      primaryClass={cs.LG},
      url={https://arxiv.org/abs/2505.23606}, 
}

@article{lladav,
  title={{LLaDA-V}: Large Language Diffusion Models with Visual Instruction Tuning},
  author={You, Zebin and Nie, Shen and Zhang, Xiaolu and Hu, Jun and Zhou, Jun and Lu, Zhiwu and Wen, Ji-Rong and Li, Chongxuan},
  journal={arXiv preprint arXiv:2505.16933},
  year={2025}
}

@article{llada1.5,
  title={{LLaDA} 1.5: Variance-Reduced Preference Optimization for Large Language Diffusion Models},
  author={Zhu, Fengqi and Wang, Rongzhen and Nie, Shen and Zhang, Xiaolu and Wu, Chunwei and Hu, Jun and Zhou, Jun and Chen, Jianfei and Lin, Yankai and Wen, Ji-Rong and Li, Chongxuan},
  journal={arXiv preprint arXiv:2505.19223},
  year={2025}
}

@inproceedings{xu2024imagereward,
  author       = {Jiazheng Xu and
                  Xiao Liu and
                  Yuchen Wu and
                  Yuxuan Tong and
                  Qinkai Li and
                  Ming Ding and
                  Jie Tang and
                  Yuxiao Dong},
  editor       = {Alice Oh and
                  Tristan Naumann and
                  Amir Globerson and
                  Kate Saenko and
                  Moritz Hardt and
                  Sergey Levine},
  title        = {ImageReward: Learning and Evaluating Human Preferences for Text-to-Image
                  Generation},
  booktitle    = nips,
  year         = {2023}
}

@article{bagel,
  title={Emerging properties in unified multimodal pretraining},
  author={Deng, Chaorui and Zhu, Deyao and Li, Kunchang and Gou, Chenhui and Li, Feng and Wang, Zeyu and Zhong, Shu and Yu, Weihao and Nie, Xiaonan and Song, Ziang and others},
  journal={arXiv preprint arXiv:2505.14683},
  year={2025}
}

@article{wu2025qwen,
  title={Qwen-image technical report},
  author={Wu, Chenfei and Li, Jiahao and Zhou, Jingren and Lin, Junyang and Gao, Kaiyuan and Yan, Kun and Yin, Sheng-ming and Bai, Shuai and Xu, Xiao and Chen, Yilei and others},
  journal={arXiv preprint arXiv:2508.02324},
  year={2025}
}
}
\end{document}